\documentclass{article}

\usepackage[final]{corl_2020} % Uncomment for the camera-ready ``final'' version.
\usepackage{graphicx}
\usepackage{comment}
\usepackage{amsmath,amssymb} %
\usepackage[table,dvipsnames]{xcolor}
\usepackage{xspace}
\usepackage{soul}
\usepackage{subcaption}
\usepackage{enumerate}
\usepackage{array}
\usepackage{xspace}
\usepackage{listings}
\usepackage{booktabs,wrapfig}
\usepackage{url}
\usepackage{xparse}
\usepackage{ifthen}
\usepackage{graphicx}
\usepackage{pdfpages}
\usepackage{appendix}
\ExplSyntaxOn
\seq_new:N \g_rolch_seq 
\NewDocumentCommand{\StoreNote}{+m}{%
  \seq_gput_right:Nx \g_rolch_seq {{#1}} % Storing all notes in a list
}
\NewDocumentCommand{\ShowNotes}{}{%
  \setcounter{footnote}{0}
  \seq_map_inline:Nn \g_rolch_seq {\stepcounter{footnote}\footnotetext{##1}}
}
\ExplSyntaxOff
\AtEndDocument{%
  \ShowNotes%
}
\newcolumntype{L}{>{\arraybackslash}m}

\newcommand{\embodiedai}{E-AI\xspace}

\newcommand{\eg}{\emph{e.g.}\xspace}

\newcommand{\openaienv}{\textsc{Env}\xspace}
\newcommand{\env}{\textsc{Environment}\xspace}
\newcommand{\task}{\textsc{Task}\xspace}
\newcommand{\tasksampler}{\textsc{TaskSampler}\xspace}

\newcommand{\experimentconfig}{\textsc{ExperimentConfig}\xspace}
\newcommand{\trainingpipeline}{\textsc{TrainingPipeline}\xspace}
\newcommand{\pipelinestage}{\textsc{PipelineStage}\xspace}

\makeatletter
\def\blfootnote{\xdef\@thefnmark{}\@footnotetext}
\makeatother

\title{AllenAct: A Framework for Embodied AI Research}

\author{
    Luca Weihs$^{*1}$, Jordi Salvador$^{*1}$, Klemen Kotar$^{*1}$, Unnat Jain$^{2}$, Kuo-Hao Zeng$^{3}$\\ \textbf{Roozbeh Mottaghi$^{1,3}$, Aniruddha Kembhavi$^{1,3}$} \\
    $^1$ PRIOR @ Allen Institute for AI \,\,\,\, $^2$ UIUC \,\,\,\, $^3$ University of Washington \\ \\
    \url{https://allenact.org}
}

\definecolor{upcomingcolor}{RGB}{122, 112, 243}
\definecolor{lightred}{rgb}{1.0, 0.4, 0.4}

\newcommand{\lib}{\mbox{AllenAct}\xspace}

\begin{document}
\maketitle

%===============================================================================

\begin{abstract}
The domain of Embodied AI, in which agents learn to complete tasks through interaction with their environment from egocentric observations, has experienced substantial growth with the advent of deep reinforcement learning and increased interest from the computer vision, NLP, and robotics communities. This growth has been facilitated by the creation of a large number of simulated environments (such as AI2-THOR, Habitat and CARLA), tasks (like point navigation, instruction following, and embodied question answering), and associated leader-boards. While this diversity has been beneficial and organic, it has also fragmented the community: a huge amount of effort is required to do something as simple as taking a model trained in one environment and testing it in another. This discourages good science. We introduce AllenAct, a modular and flexible learning framework designed with a focus on the unique requirements of Embodied AI research. AllenAct provides first-class support for a growing collection of embodied environments, tasks and algorithms, provides reproductions of state-of-the-art models and includes extensive documentation, tutorials, start-up code, and pre-trained models. We hope that our framework makes Embodied AI more accessible and encourages new researchers to join this exciting area.\blfootnote{$^*$ indicates equal contribution.}
\end{abstract}

%\keywords{Embodied AI, Reinforcement Learning, Imitation Learning, Vision, Framework}

%===============================================================================

%===============================================================================

\section{Introduction}\label{sec:intro}
%\vspace{-3mm}

\begin{figure*}
    \centering
    \includegraphics[width=33pc]{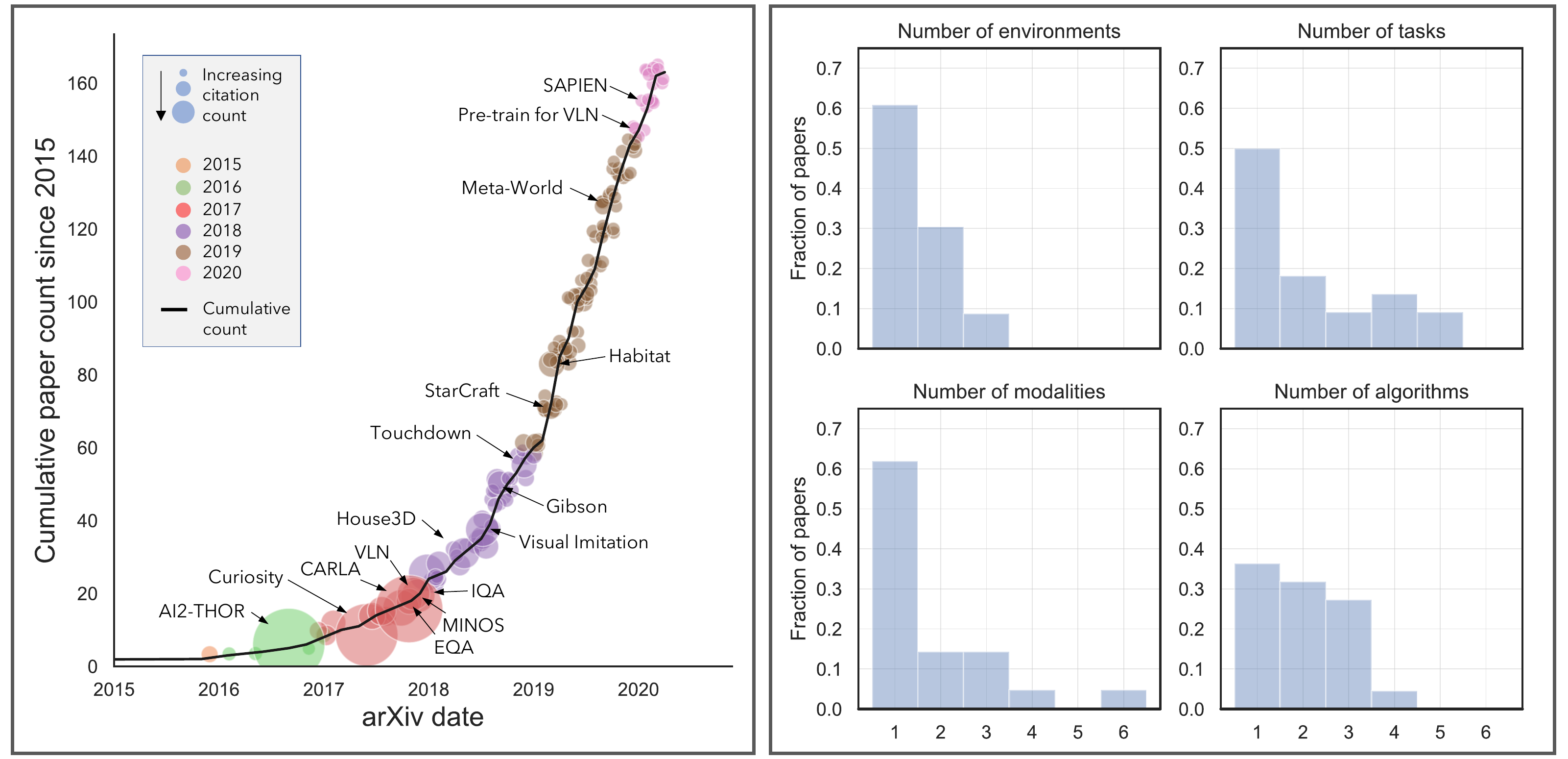}
\caption{\textbf{Growth and fragmentation of \embodiedai.} \emph{Left} - the cumulative number of papers published on arXiv since 2015 which were identified as being in the \embodiedai domain (see Appendix~\ref{app:generating-fig} for details). The number of publications has dramatically increased in recent years. \emph{Right} - after manually annotating the 20+ most-cited \embodiedai papers, we plot histograms of the frequencies with which these papers ran experiments on multiple environments, with multiple tasks, etc. The large frequency with which only a single task, environment, and modality is evaluated suggest large barriers to comprehensive evaluation. While several papers evaluate multiple algorithms, we noticed little standardization - some compare imitation with reinforcement learning, some compare A3C and PPO, others try Q-learning. Moreover, these represent the most cited papers of the past years, likely making this analysis not representative of a randomly selected paper. This analysis used the S2ORC \cite{LoWang2020s2orc}.}
    \label{fig:embodied-ai-lit}
    %\vspace{-5mm}
\end{figure*}

In recent years we have witnessed a surge of interest within the computer vision, natural language, and robotics communities towards the domain of \emph{Embodied AI} (\embodiedai) - learning, while situated within some animate body (\eg a robot), to perform tasks in environments through interaction. This has led to the development of a multitude of simulated environments employing photorealistic images (such as Gibson \cite{gibson} and AI Habitat \cite{habitat19iccv}), involving robot-object interaction (such as AI2-THOR \cite{ai2thor} and Virtual Home \cite{VirtualHome}), focused on manipulation (such as RL-Bench \cite{rlbench}, Sapien \cite{SAPIEN}, and Meta-world \cite{yu2019meta}), using advanced physics simulations (such as MuJoCo \cite{MuJoCo} and ThreeDWorld \cite{Gan2020ThreeDWorldAP}), and also physical counterparts to simulation environments (RoboTHOR \cite{robothor} and iGibson \cite{igibson}) to enable research in simulation-to-real transfer. Within these environments, research has progressed towards learning to interact: including visual navigation~\cite{yang19,ddppo,chaplot2020learning}, question answering~\cite{gordon18,embodiedqa}, task completion~\cite{Zhu2017VisualSP}, instruction following \cite{Anderson2018VisionandLanguageNI,ALFRED20}, language grounding~\cite{retouchdown, REVERIE}, 
%grasping \cite{Levine2018LearningHC, Pinto2016SupersizingSL, Gupta2018RobotLI},
grasping~\cite{Levine2018LearningHC, Gupta2018RobotLI},
object manipulation~\cite{Fan2018SURREALOR, ikea}, future prediction~\cite{khz2020visualreaction}, and multi-agent collaboration~\cite{twobodyproblem, cordialsync}; as well as using interaction as a tool to learn: environment representations~\cite{hideseek}, intuitive physics~\cite{battaglia16}, and objects and attributes~\cite{Lohmann2020LearningAO}. The rapidly growing list of publications in \embodiedai, see Fig.~\ref{fig:embodied-ai-lit}, as well as popularity of \embodiedai workshops and challenges in top computer vision and machine learning conferences over the past couple of years exemplify the growing interest in this domain.

As the domain of \embodiedai continues to grow, it faces several challenges: 
(a) \emph{Replication across tasks and datasets} - While our community proposes a host of novel methods each publication cycle, these techniques are frequently evaluated on a single task and within a single simulation environment (see Fig. \ref{fig:embodied-ai-lit}). Just as we now expect neural architectures to be evaluated across multiple tasks (\eg computer vision tasks include classification, detection, and segmentation) and multiple datasets (such as ImageNet \cite{imagenet} and Places \cite{places}), we must also start evaluating \embodiedai methods across tasks and datasets. Unfortunately, this currently requires large-scale changes to a code-base and thereby discourages comprehensive evaluation. (b) \emph{Unravelling what matters} - As the field progresses via improvements on standard tasks and benchmarks, it is crucial to understand what components of systems matter most and which do not matter at all. This unravelling requires careful ablation studies and analyses. In addition to evaluations across tasks and datasets, this involves swapping out learning algorithms (e.g., on-policy and off-policy), losses (e.g., primary and auxiliary), model components (representation stacks, maps, etc.), and hyperparameters. These analyses are often critical for good science and also fast progress. Today's frameworks and libraries can certainly be improved in this regard. Why should swapping a learning algorithm (e.g., PPO with A2C) be any more tedious than changing, \eg, the learning rate? 
(c) \emph{Ramp up time} - Getting up to speed with \embodiedai algorithms takes significantly longer than ramping up to classical tasks in vision and NLP like image classification or sentiment analysis. Just as the early deep learning libraries like Caffe and Theano, and numerous online tutorials, lowered entry barriers and ushered in a new wave of researchers towards deep learning, \embodiedai can benefit from modularized coding frameworks, comprehensive tutorials, and ample startup code. 
(d) \emph{Large training cost} - \embodiedai is expensive. Today's state of the art reinforcement learning (RL) methods are sample inefficient and training competitive models for embodied tasks can cost tens of thousands of dollars - within the reach of industrial AI labs, but unaffordable for independent researchers and smaller organizations. The availability of large networks pre-trained on ImageNet (with accompanying code and models on standard libraries like PyTorch) significantly reduced the cost of training on downstream tasks. A similar centralized repository with a diverse set of \embodiedai code and models can greatly benefit our community.

As detailed in Section \ref{sec:related-work}, there is no shortage of open-source reinforcement learning libraries and frameworks available today. While these frameworks excel in their particular domains, for research in \embodiedai we found that each individually lacked features we consider critical.  In particular, no single framework simultaneously provides: support for a large number of \embodiedai environments and tasks, a variety of training algorithms, a capacity to construct custom training pipelines, the right balance between adding new and exploiting existing functionality, and a high likelihood of continued support and development. For this reason, we set out to develop a new framework focused on \embodiedai research.

We present the \lib framework, written in Python and using PyTorch \cite{pytorch}, designed for research in \embodiedai with a focus on modularity, flexibility, and well encapsulated abstractions. It inherits the best design principles and builds upon other AI libraries including \texttt{pytorch-a2c-ppo-acktr}\footnote{\url{https://github.com/ikostrikov/pytorch-a2c-ppo-acktr-gail}} and \texttt{Habitat-API} \cite{habitat19iccv}.    %\footnote{\url{https://github.com/facebookresearch/habitat-api}}.
While \lib will continue to improve, we highlight the following existing features: (1) \emph{Environments} - we provide first-class support for the iTHOR \cite{ai2thor}, RoboTHOR \cite{robothor}, and Habitat \cite{habitat19iccv} embodied environments and numerous tasks within, as well as for grid-worlds including MiniGrid \cite{minigrid}. Grid-worlds serve as excellent sand-boxes to evaluate new algorithms owing to their rendering speed and variable complexity. Swapping out environments, as well as adding new ones, is made simple. (2) \emph{Task Abstraction} - tasks and environments are decoupled in \lib. This allows researchers to easily implement a large variety of tasks in the same environment. (3) \emph{Algorithms} - we provide support for a variety of on-policy algorithms including PPO \cite{Schulman2017ProximalPO}, DD-PPO \cite{ddppo}, A2C \cite{Mnih2016AsynchronousMF}, Imitation Learning (IL), and DAgger \cite{Ross2011ARO} as well as offline training such as offline IL. (4) \emph{Sequential Algorithms} - \lib makes it trivial to experiment with different sequences of training routines, which are often the key to successful policies (example: IL followed by PPO). (5) \emph{Simultaneous Losses} - \lib allows researchers to easily combine various losses while training models (for instance, use an external self-supervised loss while optimizing a PPO loss). While seemingly trivial, we found that present day RL libraries make this unnecessarily  harder than it need be. (6) \emph{Multi-agent support} - \lib provides support for multi-agent algorithms and tasks. (7) \emph{Visualizations} - effective visualizations of embodied environments are critical for debugging and ideation. \lib provides out-of-the-box support to easily visualize first person and third person cameras for agents as well as intermediate model tensors and integrates these into Tensorboard. (8) \emph{Pre-trained models} - \lib provides a number of models and accompanying code to train these models for standard \embodiedai tasks. (9) \emph{Tutorials} - we provide start-up code to help ramp up new researchers to the field of embodied-AI as well as tutorials for performing common actions like adding new environments, tasks, and models.

The \lib framework will be made open source and freely available under the MIT License. We welcome and encourage contributions to \lib's core functionalities as well as the addition of new environments, tasks, models, and pre-trained model weights. Our goal in releasing \lib is to make \embodiedai more accessible and encourage thorough, reproducible, research.
\section{Related Work}\label{sec:related-work}

\noindent \textbf{Embodied AI platforms.} AI research has benefited from platforms that enable agents to interact with, and obtain observations from, an environment. These platforms have been used as benchmarks to evaluate AI models on different types of tasks ranging from games \cite{ALE} to performing tasks in indoor environments \cite{ai2thor} to autonomous driving \cite{Dosovitskiy17}.
ALE \cite{ALE}, ViZDoom \cite{vizdoom}, and Malmo \cite{malmo}
are example game environments. Arena \cite{song2019arena} provides a multi-agent platform for games. Several efforts have produced environments for navigation with virtual robotic agents \cite{robothor,habitat19iccv,igibson}. AI2-THOR \cite{ai2thor}, CHAI \cite{Misra2018MappingIT}, and Virtual Home \cite{VirtualHome} are examples of platforms that go beyond navigation and enable evaluation of agents on tasks that require interaction such as applying forces and/or changing object states. Platforms such as RLBench \cite{rlbench}, Sapien \cite{SAPIEN}, and Meta-World \cite{yu2019meta} focus on manipulation tasks, while \cite{Len2010OpenGRASPAT,Kootstra2012VisGraBAB} enable studying the task of grasping. The DeepMind Control Suite \cite{Tassa2018DeepMindCS} provides a platform for continuous control tasks. CARLA \cite{Dosovitskiy17} is designed to evaluate autonomous driving capabilities. Our goal is to provide a framework with general abstractions so researchers can easily plug-in their environment of interest and begin experimentation. We provide code for integrating multiple environments and tasks (see Sec.~\ref{sec:library}). We will continue to add more ourselves and encourage other researchers to do the same.  OpenAI Gym \cite{openaigym} also provides a standard wrapper for a set of environments including the Atari games and MuJoCo control tasks. \lib differs from Gym in its abstractions, capabilities, and \embodiedai being its primary focus.

\noindent \textbf{Embodied AI and reinforcement learning libraries.} There have been several libraries developed over the years for \embodiedai, a few recent libraries that are most relevant to ours are discussed here.  \texttt{ML-Agents} \cite{Juliani2018UnityAG} enables defining new environments in the Unity game engine and provides a collection of example environments and RL algorithms. \texttt{PyRoboLearn} \cite{Delhaisse2019PyRoboLearnAP} provides a robot learning framework, where the idea is to disentangle the learning algorithms, models, robots, and their interface and to provide an abstraction for each of these components. \texttt{Habitat-API} \cite{habitat19iccv} is a modular framework for defining embodied tasks and agent configurations and training and evaluating these agents. There are also RL frameworks without an embodied focus, for example, \texttt{Garage}\footnote{Garage \url{www.github.com/rlworkgroup/garage}. Keras-RL \url{www.github.com/keras-rl/keras-rl}} (also known as \texttt{rllab}), OpenAI Gym \cite{openaigym}, Dopamine \cite{castro18dopamine}, and Keras-RL\addtocounter{footnote}{-1}\footnotemark.
Each of these libraries offers a unique feature set well-suited to a particular research, or production, workflow. In contrast to these libraries, \lib is designed to provide first-class support (\eg, including tutorials, starter-code, visualization, and pretrained models) for a wide range of \embodiedai tasks while also allowing for substantial flexibility in defining new training pipelines and integrating new environments and tasks.

\section{The \lib framework}\label{sec:library}
%\vspace{-0.2cm}
Designing software for AI tasks requires a delicate balance between the ease with which (a) new functionality can be added and (b) the existing functionality can be exploited. For instance, a framework designed only to train GRU based agents with the PPO algorithm to complete a navigation task within the AI2-THOR environment can narrow its API so that a user needs only to specify a small set of relevant hyperparameters before running a new experiment. This makes research within this domain extremely streamlined at the expense of flexibility: if a user now wants to try something beyond the scope of the design (\eg train with the A2C loss) they will need to dive into the internals of the framework to understand what, often substantial, changes must be made. In our experience with \embodiedai research, the frequency with which we have had to modify our software to adapt to new experimental requirements has followed the following approximate pattern:

\noindent\textbf{Daily-Weekly:} Modify hyperparameters associated with the training loss (\eg reward discount factor $\gamma$), model (\eg RNN hidden state size), optimization (\eg learning rate, batch size), and hardware (\eg number of GPUs and training processes). \\
\noindent\textbf{Weekly-Monthly:} Modify model architectures, training strategies (\eg warm-start a model with IL before training with PPO), sensor modalities (\eg adding depth maps as input to the model). \\
\noindent\textbf{Quarterly-Yearly:} Adding new environments (\eg SAPIEN), new tasks (\eg a language and vision task such as ALFRED \cite{ALFRED20}), changes to the definition of an existing task (\eg success in object-navigation requires an explicit stop signal in addition to proximity to the object), new losses to be used during training (\eg auxiliary self-supervision), and incorporating new training paradigms (\eg moving from asynchronous methods, \eg A3C \cite{Mnih2016AsynchronousMF}, to synchronous methods, \eg PPO \cite{Schulman2017ProximalPO}).

In designing \lib, we have stressed modularity and flexibility while keeping the above in mind. Thus changing hyperparameters or model architectures is trivial and making more substantial changes, such as adding a new training paradigm (\eg~deep-Q learning), requires more knowledge of the framework's internals but is still relatively straightforward. Following community standards, \lib is written in the Python programming language and heavily leverages the PyTorch library for designing deep-neural models and enabling their optimization. Next, we describe \lib's API, features, documentation, and associated pre-trained \embodiedai models.

\subsection{Abstractions and API}

The API of \lib is defined by a collection of abstractions (each corresponding to a Python class) which, themselves, are best understood in context of their relationships. At a high level, an agent is defined by an \textsc{ActorCriticModel}, observes the world using \textsc{Sensor}s, and interacts with its \env to complete a \task which defines rewards and success criteria. New instances of a \task{} (\eg navigation starting from a different point) are created, sequentially, for the agent by a \tasksampler and, during training, the agent's parameters are updated to minimize some collection of \textsc{Loss}es. Which \textsc{Loss}es are used at a particular point in training is determined by the \trainingpipeline. Rather than describing all of these abstractions in detail\footnote{See \lib's documentation for these comprehensive details.} we instead highlight how these abstractions differ from those used in most RL libraries. Our code adapts and generalizes several abstractions from Habitat-API. For instance, their \textsc{Dataset} is generalized into our \tasksampler, and while we both share a \task abstraction, theirs is used within an Open AI gym \openaienv class while ours acts as an intermediary between the agent and the environment.

\noindent\textbf{Experiments defined in code.} 
% There is an abstract class \experimentconfig to define experiment configurations. 
In \lib, experiments are defined by creating an implementation of the abstract \experimentconfig class. 
Changing hyperparameters in such files is just as simple as doing so within text-based configuration files (a necessity, as noted above, as these types of changes occur daily to weekly) but with the added benefit that, at the cost of some additional boilerplate, it is trivial to add new hyperparameters, update model architectures, etc. Moreover, writing configuration in code allows easy access to a wide range of productivity features provided by modern integrated development environments such as auto-completion and type hints. This hugely simplifies daily-weekly modifications and enables researchers to easily run several experiments. An example of how one might create an \experimentconfig implementation to train a navigation model in AI2-THOR can be seen in the documentation.%\footnote{\url{ http://allenact.org/tutorials/training-a-pointnav-model}}

\noindent\textbf{Flexible training pipelines.} Training high-quality agents often requires a pipelined approach where, for example, an agent's policy is given a warm-start by first training with IL after which reinforcement learning is used to further improve performance and generalization. While such training pipelines can be accomplished manually, \lib introduces a \trainingpipeline class which makes the concept of a training pipeline a core concept within the framework. A \trainingpipeline is defined by a collection of sequential \pipelinestage{}s which define: (a) the losses to be used, (b) the length of training and any early stopping criteria, and (c) whether or not to apply teacher forcing (see Sec. \ref{sec:algorithms}). During training, \lib moves through these stages and updates the agent accordingly. With this design, adding an IL warm-start to an experiment requires adding a single additional line of code. Thus the weekly-monthly change in training pipeline takes, at most, a few minutes and requires little additional bookkeeping.

\noindent\textbf{Decoupling the environment from the task.} A standard abstraction used within multiple RL frameworks is OpenAI Gym's \openaienv. This \openaienv class defines (i) how an agent interacts with the environment, (ii) whether or not success criteria are met, (iii) the rewards returned after every action, (iv) observations available to the agent, and (v) how to reset itself. This abstraction is an excellent fit for many settings, especially those in which the environment (\eg an Atari game) is intimately tied to the agent's intended goal (\eg beating the game). This abstraction is less natural in the setting of \embodiedai where the environment (\eg AI2-THOR, Habitat, ThreeDWorld, etc.) has no innate goal and, in fact, a huge variety of distinct goals can be defined. Within the AI2-THOR environment alone, we are aware of nine unique tasks defined by various authors ranging from navigation \cite{robothor} to multi-agent furniture moving \cite{cordialsync}. 
Instead, in \lib we disentangle the \task from the \env. The \env provides a means by which to modify environment state while the \task encapsulates the agent's goal and acts as an intermediary between the agent and the environment. The \task defines the actions available to the agent, any success criteria, and the rewards returned to the agent. Once a \task has been completed by the agent it is simply thrown away and, as described below, a new task is generated by the \tasksampler. Beyond being conceptually appealing, this decoupling can make the quarterly-yearly updates to (and additions of) tasks far easier as, generally, changes are confined to the \task class and large portions of code require no changes. This decoupling also simplifies the process of introducing new environments within the \lib framework. 

\noindent\textbf{Flexible task initialization.} As previously noted, an OpenAI Gym's \openaienv instance must be able to reset itself and so any implementation of \openaienv implicitly defines the stream of goals that the agent observes during training. This is well-suited for most RL research but is not a good fit for \embodiedai where one often needs more control over which goals are presented to the agent and their order. We instead use a \tasksampler abstraction to allow complete control of how new instances of a task are, sequentially, generated for the agent. With this abstraction, enabling curriculum learning is as simple as defining a \tasksampler that progressively samples more difficult instances of a task. \tasksampler{}s enable quick experimentation with new training strategies which we require at the weekly-monthly frequency.

Together these changes allow for considerable flexibility and provide a useful mindset by which to approach \embodiedai problems. These abstractions are sufficiently general to be of use even beyond \embodiedai research, indeed \lib has been used in a grid-world-based study of reinforcement learning methodology \cite{advisor}.
\vspace{-0.2cm}
\subsection{Features}
\vspace{-0.2cm}
\noindent\textbf{Environments and Tasks.} A key goal of \lib is to provide first-class support for a diverse range of embodied environments and tasks. In this early release, we provide support for Habitat, iTHOR, and RoboTHOR and tasks within them (See Table~\ref{tab:supported-envs}). We also provide support for MiniGrid that serves as a fast sand-box for algorithm development. In future releases, we will extend support to the recently released SAPIEN \cite{SAPIEN} and ThreeDWorld \cite{Gan2020ThreeDWorldAP} environments and associated tasks (\eg robotic manipulation). A crucial advantage of \lib is the ease at which one may test the same model (or training pipeline, loss, etc.) across multiple environments and tasks. The \lib documentation shows an example of the few changes required to an \experimentconfig to switch from one task to another in iTHOR and then move to the Habitat environment. %\footnote{\url{http://allenact.org/tutorials/transfering-to-a-different-environment-framework}} 

%%%%
\begin{wraptable}{r}{9cm}
    %\vspace{-5mm}
    \footnotesize
    \begin{tabular}{rl}\toprule
        Environment & Tasks \\ \toprule
        iTHOR \cite{ai2thor} & {\small{PointNav~\cite{Anderson2018OnEO}, ObjectNav~\cite{yang19}, ALFRED~\cite{ALFRED20}}} \\
         & \color{upcomingcolor} {FurnLift~\cite{twobodyproblem}},  \color{upcomingcolor}{FurnMove~\cite{cordialsync}} \\\midrule
        RoboTHOR \cite{robothor} & PointNav~\cite{Anderson2018OnEO}, ObjectNav~\cite{yang19} \\\midrule
        Habitat \cite{habitat19iccv} & PointNav~\cite{Anderson2018OnEO}, ObjectNav~\cite{yang19}, \color{upcomingcolor}VLN~\cite{Anderson2018VisionandLanguageNI} \\ \midrule
        MiniGrid \cite{minigrid,babyai_iclr19} & All\footnotemark \\ 
        \bottomrule
    \end{tabular}
    \caption{Environments and tasks supported in \lib. This support includes starter-code, visualization tools, and pre-trained models. Tasks and environments in {\color{upcomingcolor}purple} have support planned for upcoming releases.}
    \label{tab:supported-envs}
    %\vspace{-0.2cm}
\end{wraptable} 
%%%% An ugly hack to get the footnotes in the above table to work.
%\addtocounter{footnote}{-1}
\footnotetext{As an easily composable grid-world, MiniGrid can be used to quickly generate a huge number of different tasks. Taken together, MiniGrid and BabyAI have $\geq$30 unique tasks by default.}

\noindent\textbf{Algorithms.}\label{sec:algorithms}
Our framework currently supports decentralized, distributed, synchronous, on-policy training with A2C, PPO, IL, or any user-defined loss. This builds upon the decentralized distributed proximal policy optimization (DD-PPO) \cite{ddppo} algorithm available in the Habitat-API with the additional flexibility that training can be done with arbitrary loss functions so that we can, for example, run DD-A2C or DD-curiosity-driven-exploration \cite{Pathak2017CuriosityDrivenEB}. While on-policy and synchronous RL algorithms have become very popular, with PPO the de-facto standard, and have been used with great success, on-policy methods are notoriously sample-inefficient and synchronous methods impose run-time limitations that are not appropriate for every problem. To this end, \lib currently supports two means by which to relax the on-policy assumption. \emph{Teacher forcing} - in order to implement algorithms such as DAgger \cite{Ross2011ARO}, an agent must be able to replace its action with the action of an expert with some probability (potentially decaying over training). In \lib, implementing teacher forcing (and thus DAgger) is as simple as defining a linear-decay function. \emph{Training with a fixed, external, dataset} - it is frequently beneficial to be able to supervise an agent using a fixed dataset (e.g. IL from a dataset of human examples). \lib enables this type of supervision and also interleaving off-policy updates with on-policy ones.
While we have found the above relaxations of the synchronous and on-policy assumptions to be sufficient for most prior work, we recognize that this will not be the case for all users of \lib. In light of this, our future roadmap includes incorporating deep Q-learning methods as well as capabilities for asynchronous execution and training.

\noindent\textbf{Multiple Agents.} A key facet of \embodiedai having received relatively little attention is collaboration and communication among multiple agents. To support research in this direction we have natively enabled training multi-agent systems in \lib. As seen in Table \ref{tab:supported-envs}, we will soon provide high-quality support for the multi-agent tasks recently developed for AI2-THOR \cite{twobodyproblem,cordialsync}.

\begin{figure}
    \centering
    % {
    %   \phantomsubcaption\label{fig:}
    % }
    % \includegraphics[width=\textwidth,height=3cm]{example-image-b}
    \includegraphics[width=\textwidth]{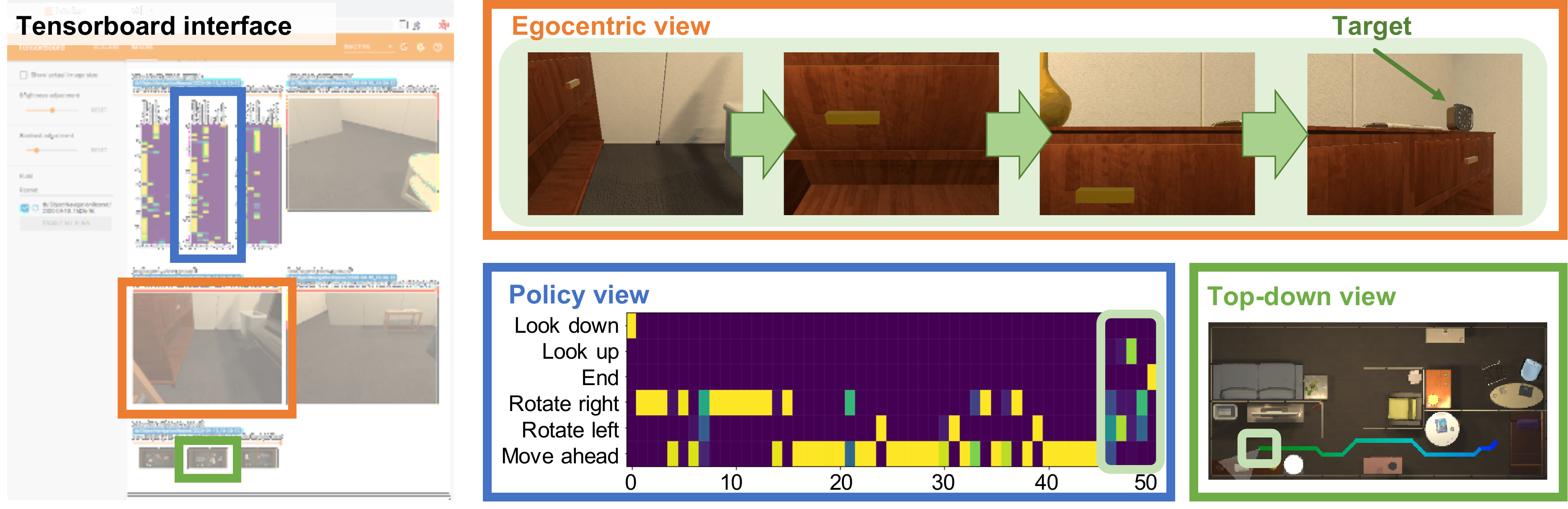}
    \caption{\textbf{Visualizations.} A simple plug-in-based interface allows generating different types of visualizations in Tensorboard for a particular task as shown in the screen capture on the left. A navigation task is shown in this example. We display ego-centric and top-down views as well as the policy over time. The top-down view enables observing the overall behavior of the agent. We highlight one segment of the trajectory with a green overlay. The ego-centric view enables interpreting the policy and visually assessing task success. The policy enables analyzing and debugging the probability of each action at each point in time.}
    \label{fig:visualizations}
    \vspace{-0.2cm}
\end{figure}

\noindent\textbf{Visualization.} In our experience, visualizations and qualitative evaluations of the policies learned by \embodiedai agents are critical to debugging and understanding the limitations of current systems. Unfortunately, producing such visualizations can be time consuming, especially so if these visualizations are meant to be of sufficient quality to be used in presentations or publications. To lower the burden of visualization, the \lib framework contains a number of utilities (including ego-centric views, top-down views, third party views and tensor visualizations) for environments with first-class support (recall Table \ref{tab:supported-envs}). Some of these visualizations, which can be automatically logged during inference, are presented in Figure~\ref{fig:visualizations}. The range and scope of these visualization utilities will grow as further embodied environments and tasks are incorporated into our framework.

\noindent\textbf{Tutorials, Documentation, and Typing.} Beginning to work with a new framework can be a daunting and frustrating experience as one must internalize a large number of new abstractions, frequently with little documentation, written in an unfamiliar style. For this reason, we have made tutorials and documentation a high priority within \lib and, in our first release, we have several tutorials such as training a PointNav model in RoboTHOR or how to switch environments for a particular task. Moreover, we have added type hints throughout our codebase to enable IDE auto-completion and warnings.

\noindent\textbf{Pre-trained models.}
To encourage reproducibility, we include several pre-trained model checkpoints (reproducing, within error, published results) for tasks with first-class support. This includes all models trained towards the experiments in Sec.~\ref{sec:experimental-results}. As these models were trained within \lib, we also provide training and inference code for these models.

\noindent\textbf{Future development and support.}
An important consideration when deciding whether or not to adopt a new framework is that of future support. Our team is committed to research in the domain of \embodiedai and expect to continue \lib's development for, at least, several more years. Indeed we currently have a number of ongoing projects using \lib, in executing these projects we expect to obtain robust feedback that will be used to improve \lib. We will also encourage and make it easy for \embodiedai researchers to contribute code and models to \lib.
\vspace{-0.2cm}
\section{Experiments}\label{sec:experimental-results}
\vspace{-0.2cm}
We now highlight the capabilities of \lib by reproducing results from the (embodied) RL literature along with evocative ablations. Code and model checkpoints for all experiments can be easily accessed within \lib and serve as strong comparative baselines for future work.

\begin{figure}[t!]
    \vspace{-0.5cm}
    \centering
    {
      \phantomsubcaption\label{fig:nav-results}
      \phantomsubcaption\label{fig:babyai-gotolocal-results}
      \phantomsubcaption\label{fig:minigrid-advisor-results}
    }
    \begingroup
    \setlength{\tabcolsep}{1pt} % Default value: 6pt
    \renewcommand{\arraystretch}{0.1} % Default value: 1
    \tiny{
    \begin{tabular}{cccc}
        \includegraphics[height=0.31\textwidth]{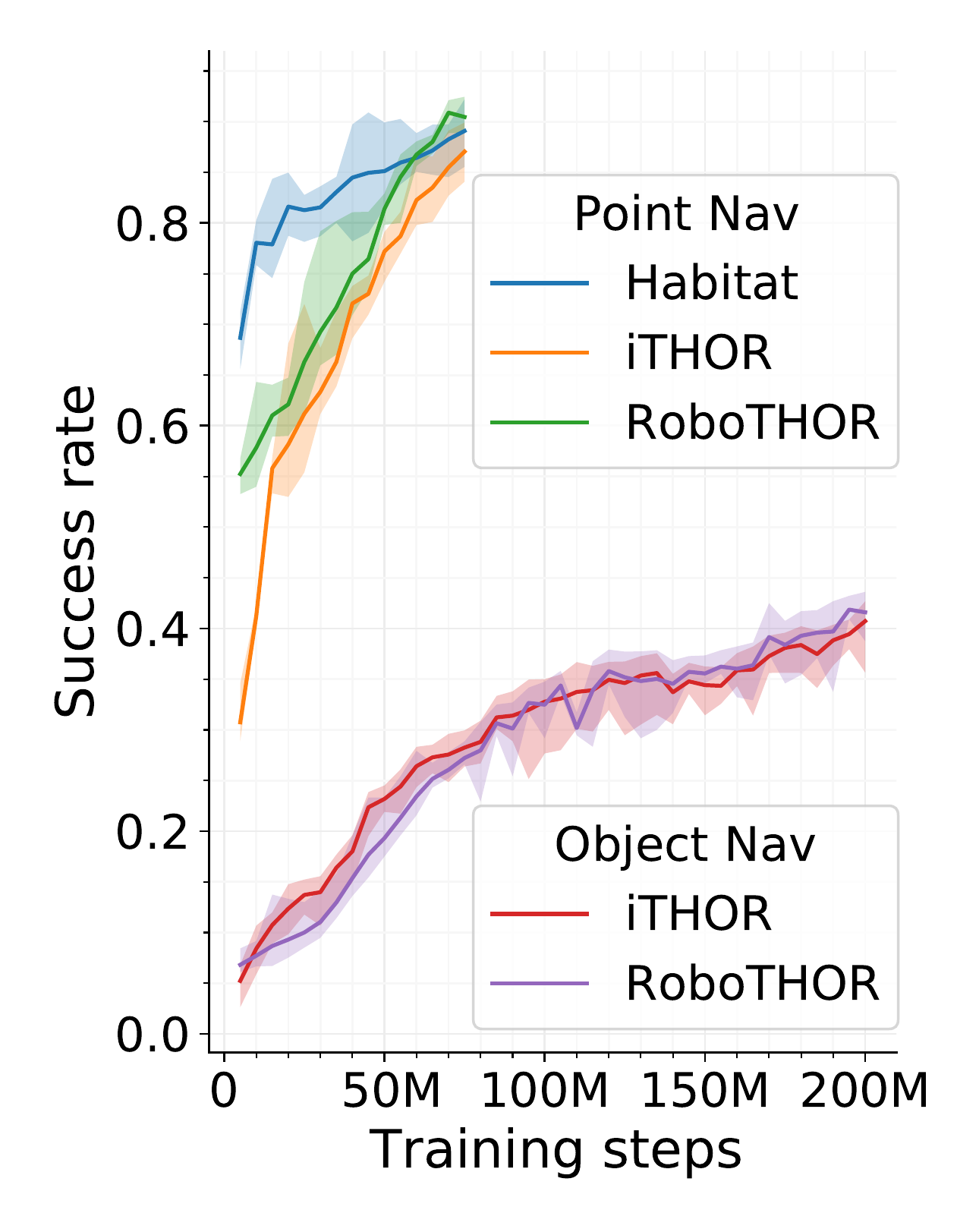} &
        \includegraphics[height=0.31\textwidth]{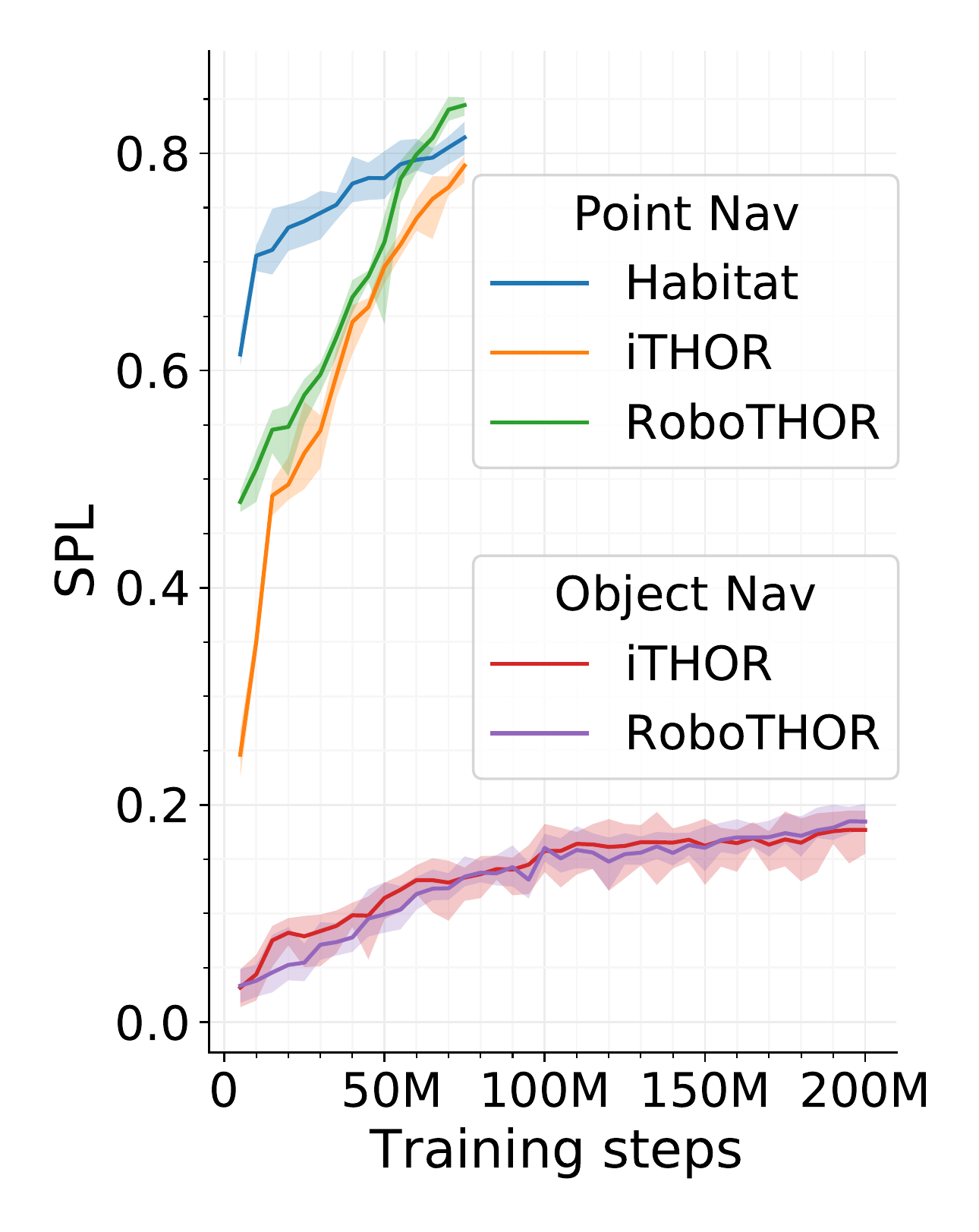}  & 
        \includegraphics[height=0.31\textwidth]{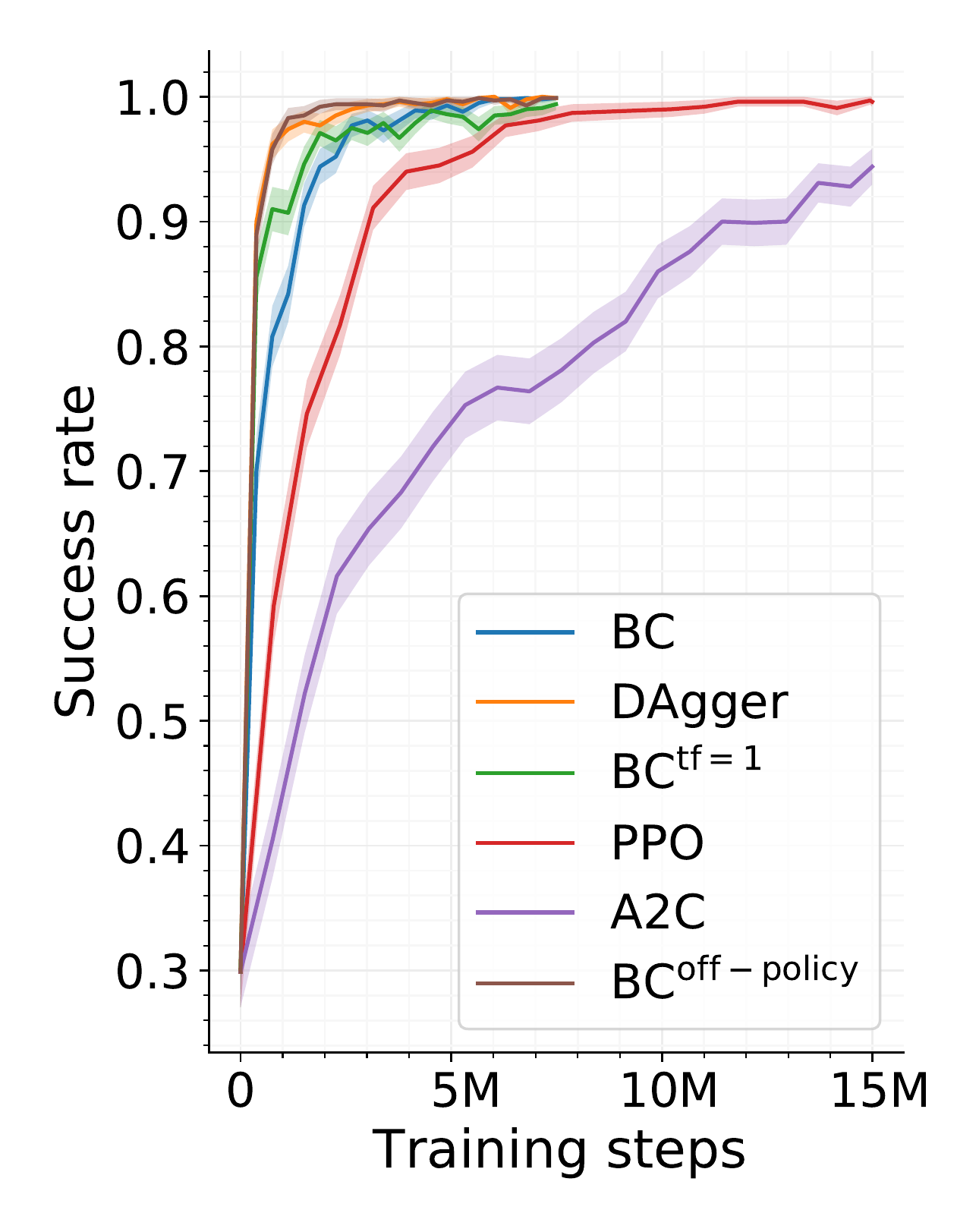} &
        \includegraphics[height=0.31\textwidth]{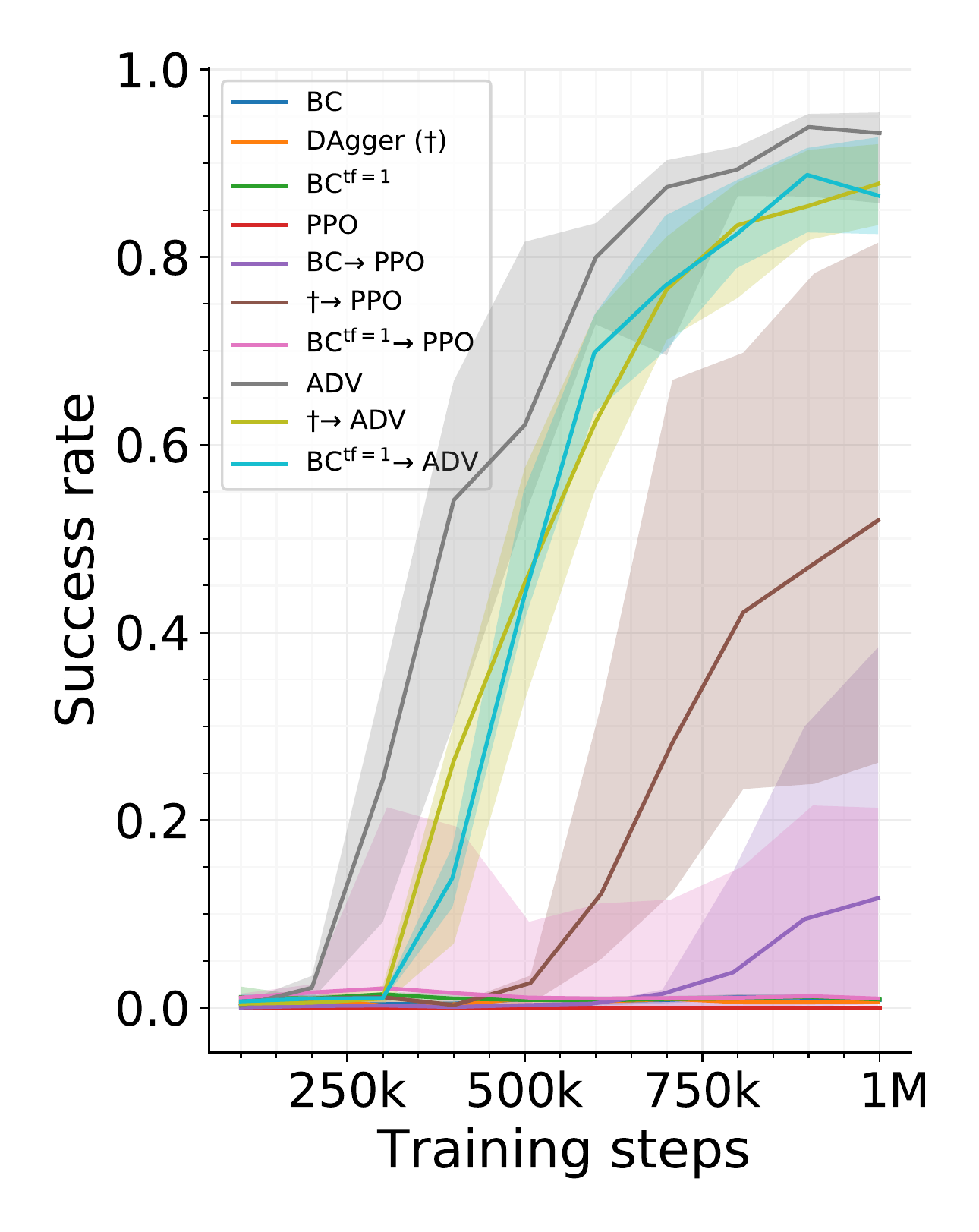} \\
         \multicolumn{2}{c}{\textbf{(a)} PointNav and ObjectNav (RoboTHOR, iTHOR, Habitat)} &
         \textbf{(b)} GoToLocal (BabyAI) & \textbf{(c)} LC Corrupt S15N7 (MiniGrid)\\
    \end{tabular}
    }
    \endgroup
    \caption{\textbf{Experimental results.} \textbf{(a)} Validation-set success rate and Success weighted by Path Length (SPL) \cite{Anderson2018OnEO} of models trained with DD-PPO to complete the PointNav \& ObjectNav tasks in iTHOR, RoboTHOR \& Habitat. 
     Shaded regions indicate the range of metrics in five runs for PointNav in Habitat, three runs elsewhere. \textbf{(b)} Test-set success rate of models trained to complete BabyAI's GoToLocal task (shaded regions indicate 95\% confidence intervals around the mean); \textbf{(c)} Test-set success rate of models trained to complete the LC Corrupt S15N7 from \cite{advisor} (shaded regions indicate inter-quartile ranges across 10 runs).}
    \label{fig:exp_results}
    \vspace{-0.4cm}
\end{figure}

\noindent \textbf{Support for embodied environments and tasks.}
Using \lib, we have reproduced a number of results for navigation in the iTHOR, RoboTHOR, and Habitat environments, see Fig. \ref{fig:nav-results}. Two variants of navigation are commonly considered in the literature: PointNav (navigating with a displacement vector to the goal) and ObjectNav (navigating to goal specified by a category label). We train DD-PPO~\cite{ddppo} for $75$ million frames and obtain a validation accuracy of 92.5\%. This accuracy is within error of, and indeed slightly outperforming, the model in \cite{habitat19iccv}. \cite{ddppo} demonstrated that if trained for 2.5 billion frames using $\approx$4608 GPU hours DD-PPO can reach 99.9\% validation set accuracy. For our aim of demonstrating reproducibility and functionality, we restrict ourselves to 75 million frames, the same as~\cite{habitat19iccv} with whom we compare. When training to complete PointNav in iTHOR and RoboTHOR we obtain similarly high performance.

We demonstrate the ObjectNav task on iTHOR and RoboTHOR. In these experiments, we use a ResNet and LSTM based architecture and train our models for 200 million steps. Within RoboTHOR, our model outperforms the best model submitted to the recent CVPR'20 RoboTHOR challenge\footnote{\url{https://ai2thor.allenai.org/robothor/challenge/}}.
% and thus achieves state-of-the-art results. 
While no such challenge was undertaken for iTHOR, our similarly high metrics suggest we have obtained a well-trained model that will serve as a strong baseline for future work. Implementation details are in Appendix~\ref{app:support-for-embodied-envs}.

\noindent \textbf{Support for online and offline algorithms.}\label{sec:babyai-experiments}
\lib supports a range of different built-in training algorithms beyond PPO including A2C and several varieties of IL (\eg on-policy behavior cloning, DAgger, and purely offline training from a fixed dataset of demonstrations). Precise details of these IL-based methods are given in Appendix~\ref{sec:exp-online-offline}. In Fig.~\ref{fig:babyai-gotolocal-results}, we highlight the results of using these algorithms to train an agent to complete the GoToLocal task in the BabyAI \cite{babyai_iclr19} grid-world environment. Reproducing similar results as Chevalier-Boisvert et al. \cite{babyai_iclr19}, we find that IL and RL-based algorithms can be used to train agents to nearly perfect test-set accuracy within a few million steps but that IL methods converge far faster in general. While Chevalier-Boisvert et al. trained their RL-based models only with PPO, our experiments suggest that A2C can be effective but appears to be substantially less sample efficient for this task.

\noindent \textbf{Support for simultaneous and sequential algorithms.}
Beyond offering popular online and offline algorithms, \lib facilitates training with multiple losses in sequence (using a \trainingpipeline) or in parallel. Sequential training, particularly $\text{IL}\to\text{RL}$, is widely adopted in \embodiedai to \textit{warm-start} agents~\cite{Zhu2017VisualSP,embodiedqa,twobodyproblem} and has recently been formally studied~\cite{advisor}. Reproducing similar results as~\cite{advisor}, we find that a $\text{IL}\to\text{RL}$ combination of $\text{BC}$, $\text{DAgger}$, or $\text{BC}^{\text{tf=1}}$ (as variants of $\text{IL}$) followed by $\text{PPO}$ can significantly boost their individual (near-zero) performance (see Fig.~\ref{fig:minigrid-advisor-results}). Moreover ADV, an adaptive, parallel, combination of IL and RL losses, performs the best on the challenging, MiniGrid-based, \textsc{LC Corrupt} task~\cite{advisor} (see Appendix~\ref{app:sim_seq_alg} for details). In line with \cite{advisor}, we choose hyperparamers via random search with $50$ samples for each baseline.
For each baseline, the best hyperparameter combination (out of the $50$) is selected and we train $10$ models with different seeds. We summarize the test-set performance across these runs
%in Fig.~\ref{fig:exp_results}~b)
by plotting their medians and inter-quartile ranges.

\noindent \textbf{Support for multi-agent systems.}
We reproduce the Markov Stag Hunt~\cite{peysakhovich2017prosocial}.
In our re-implementation, two agents navigate in a $5\times5$ map with $1$ Stag and $2$ Plants, for $45$ steps. 
Agents can move in any of the four cardinal directions to goal of collecting a Stag or Plant. An agent gains $+1$ reward when co-located with a Plant but can obtain a much larger reward ($+5$) if it coordinates with the other agent so that they occupy the same location as the Stag simultaneously. Alternatively, the agent receives a $-g$ penalty when it is co-located with the Stag but the other agent is not.
A Stag or Plant disappears when collected and then re-spawns randomly. As in~\cite{peysakhovich2017prosocial}, the Stag moves towards to the closet agent at each step. 
Our reproduced model achieves $124.7\pm{2.47}$, and $57.0\pm{4.21}$ reward as $g=\{0.5, 3.0\}$ at $3M$ training steps (roughly $66.7K$ training episodes).
A similar baseline model converges to $\approx120$ reward at roughly $90K$ training episodes in~\cite{peysakhovich2017prosocial}. As seen in prior work, qualitatively different behavior emerges depending on the choice of $g$ (details in Appendix~\ref{app:multi_agent}).

\noindent \textbf{Support for vision-language-embodiment.} We re-implement ALFRED \cite{ALFRED20}, which is an embodied language instruction following task. We are able to reproduce the results in the paper. More specifically, using their pre-trained model, we obtain a 4.0\% task success rate and 9.4\% on the goal condition success rate on the test-seen scenario.
\section{Conclusion}\label{sec:conclusion}
\vspace{-0.3cm}
We present \lib, a framework for reproducible and reusable research in the \embodiedai domain. Our framework provides a high degree of support (in the form of pre-trained models, starter code, and tutorials) for a growing collection of \embodiedai and general RL environments and tasks. 
%===============================================================================

% The maximum paper length is 8 pages excluding references and acknowledgements, and 10 pages including references and acknowledgements

%\clearpage
% The acknowledgments are automatically included only in the final version of the paper.
%\acknowledgments{}

%===============================================================================

%\footnotesize
% \bibliography{bibliography_arxiv}
%}  % .bib

%\newpage

% THIS IS DONE TO LINE NUMBERS MATCH IN SUPPLEMENT
% ~\\~\\~\\~\\~\\~\\~\\~\\~\\~\\~\\~\\~\\~\\~\\~\\~\\~\\~\\~\\~\\~\\~\\~\\~\\~\\~\\~\\~\\~\\~\\~\\~\\~\\~\\~\\~\\~\\~\\~\\~\\~\\~\\~\\~\\~\\~\\~\\~\\~\\~\\~\\~\\~\\~\\~\\~\\~\\~\\~\\~\\~\\~\\~\\~\\~\\~\\~\\~\\~\\~\\~\\~\\~\\~\\~\\~\\~\\~\\~\\~\\~\\~\\~\\
% \newpage

%===============================================================================
\appendix

% \section*{Supplementary Materials for ``A Framework for Reproducible, Reusable, and Robust \lib Research''}

% Our supplementary appendices are organized as follows:
% \begin{enumerate}
%     \item[App. \ref{app:generating-fig}] -- A discussion of the data used in generating Fig. \ref{fig:embodied-ai-lit}.
%     \item[App. \ref{app:experiment-details}] -- Additional details for our experimental results.
%     \item[App. \ref{app:tutorials}] -- Three examples of tutorials included in our framework.
% \end{enumerate}

% Also included in our supplementary files are: (a) the \lib codebase, (b) the \lib documentation, and (c) a \texttt{DOCS-README.txt} file giving additional information on how to host the \lib documentation on your local machine.
~\\~\\
\noindent{\Large{\textbf{Appendix}}}

\section{Generating Figure \ref{fig:embodied-ai-lit}} \label{app:generating-fig}

Citation counts for papers in Fig~\ref{fig:embodied-ai-lit}-left were obtained using The Semantic Scholar Open Research Corpus (S2ORC) \cite{LoWang2020s2orc}. We restricted this analysis to papers submitted on arXiv within the past 6 years (2015 to 2020). Papers were determined to being in the \embodiedai domain if either the title or abstract contained at least 1 word from set A and at least 1 word from set B, shown below: 

\noindent \textbf{Set A:} \texttt{habitat, gibson, igibson, ai2-thor, ai2thor, matterport, matter-port, matter-port3d, matterport3d, r2r, room-to-room, house3d, pybullet, vizdoom, chai, malmo, rlbench, procgen, touchdown, retouchdown, carla, minos, chalet, meta-world, sapien, mujoco, replica, house3d, embodiedqa}

\noindent \textbf{Set B:} \texttt{robot, agent, embodied, simulator, autonomous}

The annotations for the histograms in Fig~\ref{fig:embodied-ai-lit}-right were produced manually by the authors of this submission. The top 23 papers (sorted by S2ORC citation counts) were used for this analysis.

\section{Experiment Details} \label{app:experiment-details}
In this section we provide additional details about the experiments listed in the main paper.

\subsection{Support for embodied environments and tasks.} \label{app:support-for-embodied-envs}
\noindent\textbf{Methods.} 
We trained all the models using DD-PPO. We trained the ObjectNav models for 200M steps and the PointNav models for 75M steps. For all the models we used a starting learning rate of 3e-4 and have annealed it linearly to 0, over the course of the training run. We used rollout lengths of 30 and a $\gamma$ of 0.99. \\

\noindent\textbf{Task.} 
We defined success on the PointNav task as taking the stop action within 0.2m from the target. We define success on the ObjectNav task as taking the stop action while looking at the target at a distance of no more than 1.0m. We used a turning angle of 30 degrees and a forward motion distance of 0.25m for all the experiments.

\subsection{Support for online and offline algorithms.} \label{sec:exp-online-offline}

In Section \ref{sec:babyai-experiments} we trained a number of different RL and IL baseline methods to complete the GoToLocal task in the BabyAI environment. We describe the details of these baseline methods, along with relevant hyperparameters below. For all of these experiments we use the same model used by Chevalier-Boisvert et al. \cite{babyai_iclr19}.
% The training loop in our framework iteratively repeats the following steps until completion:
% \begin{enumerate}[1.]
%     \item For a rollout 
% \end{enumerate}

\noindent\textbf{Methods.}
\begin{itemize}
    \item \emph{PPO} \cite{Schulman2017ProximalPO} -- proximal policy optimization is an onpolicy, synchronous, RL algorithm which uses an easy-to-implement clipping methodology allowing for multiple gradient-updates with a single collection of rollouts from the agent's policy to obtain better sample efficiency than several other popular approaches. 
    % \jordi{it's not really about clipping, right? what about the use of prob ratios instead of logprobs in reinforce or a2c?}
    For each update we use 1,536 rollouts of length 32 which are broken into batches of size $384{\times}32$ and iterated across 4 times (a total of 16 gradient updates per collection of rollouts). We use a fixed learning rate of $10^{-4}$, a reward discount factor of $\gamma=0.99$ and a clipping parameter of $0.1$ (linearly decaying to $0$ over training). Further training details can be found in our code base.
    
    \item \emph{A2C} \cite{Wu2017ScalableTM} -- advantage actor critic is a synchronous variant of A3C (\cite{Mnih2016AsynchronousMF}) which, empirically, often results in better performance. While A2C has fallen out of favor, with PPO largely taking its place as the de facto onpolicy, synchronous, RL algorithm, it remains a strong comparative baseline. For each update we use 768 rollouts of length 16. With A2C, each such sample is used for only a single gradient update and the rollouts are together a single batch (this is not a limitation of \lib, we wished instead to remain faithful to the standard implementation of A2C). As A2C's hyperparameters are a subset of those of PPO, we used the same hyperparameters as from PPO when applicable.
    
    \item \emph{BC} - behavioral cloning is a straightforward variant of imitation learning in which: (i) rollouts are collected using the agent's current policy, (ii) for each such action, we compute and store the, possibly different, expert's action, and (iii) the agent's policy is trained by minimizing a negative cross entropy loss between the agent's policy and the expert action. For each such update we used 128 rollouts each of length 128. Similarly as for PPO, we use these rollouts to create 4 batches of size 32 and iterate over these batches 4 times for a total of 16 gradient updates per collection of rollouts. We use a learning rate of $10^{-3}$ which decays linearly to 0 throughout training.
    
    \item \emph{DAgger} - training of DAgger is essentially identical to BC except where instead of always sampling actions from the agent's policy we use the expert's action with probability starting at 1 and decaying linearly to 0 during training. This means that the agent will initially see many successful rollouts which can improve training performance.
    
    \item \emph{BC$^{\text{tf=1}}$} - in this variant of behavioral cloning we \emph{always} take the expert's action. This is equivalent to training using a fixed dataset of expert trajectories but where this dataset is sufficiently large that no trajectory is seen more than once during training. All other hyperparameters are identical to as in BC.
    
    \item \emph{BC$^{\text{off-policy}}$} - this is imitation learning using a fixed dataset of 1 million expert demonstrations. Unlike in BC$^{\text{tf=1}}$, the agent will see the expert trajectory multiple times throughout training. While the training batches used in BC$^{\text{off-policy}}$ are of the same size as in the above IL methods, a single trajectory is not iterated over more than once until an entire epoch over the dataset is complete. In our experience this difference generally means that BC$^{\text{off-policy}}$ will obtain better results early in training (when compared to BC$^{\text{tf=1}}$) but, as BC$^{\text{tf=1}}$ is not limited to to a fixed number of experiences, with BC$^{\text{tf=1}}$ there is no fixed training set to which it can overfit.
\end{itemize}

\begin{figure} [tp]
    \centering
    \includegraphics[width=0.80\textwidth]{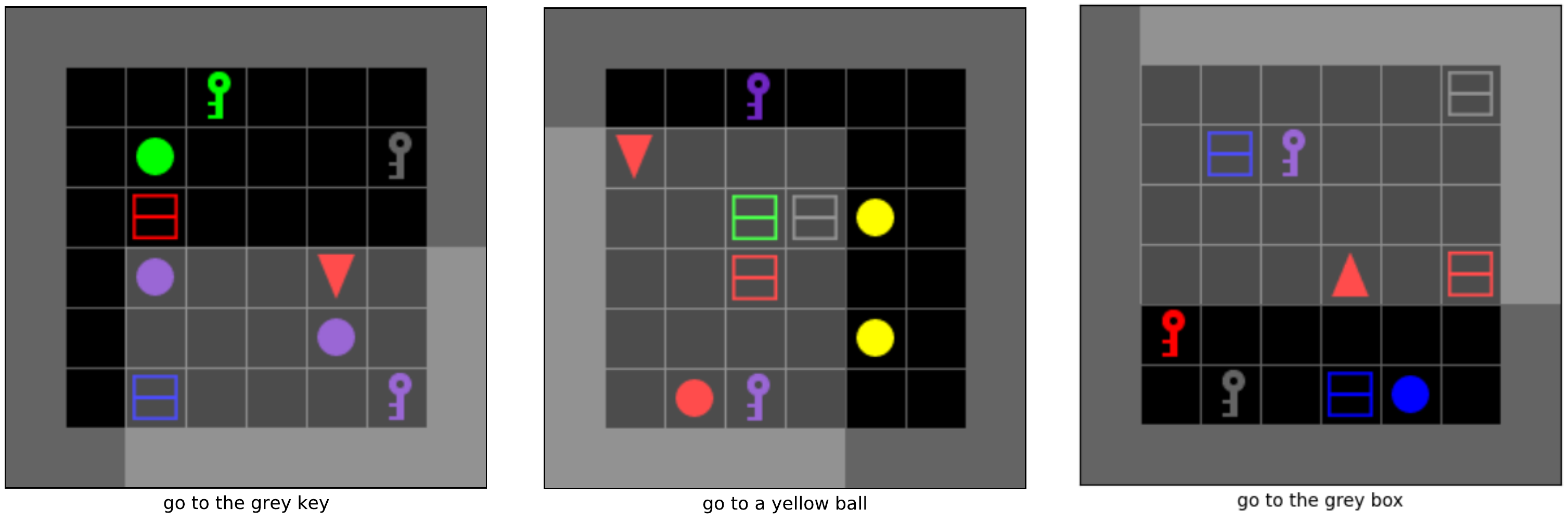}
\caption{\textbf{BabyAI's GoToLocal.} Three visualizations of the GoToLocal task in the BabyAI environment. At every step the agent (red triangle) obtains a $7{\times}7$ egocentric observation corresponding to the highlighted region and must follow a given ``go to'' instruction. E.g. in the leftmost image, the agent must ``go to the grey key''.}
    \label{fig:gotolocal-example}
\end{figure}

\noindent\textbf{Task.} Some examples of the GoToLocal task are given in Fig. \ref{fig:gotolocal-example}. At each step the agent is given a $7{\times}7$ egocentric observation and a five-word instruction defining the object to which it must navigate. %\jordi{if it's GoToLocal why is the target an object?}

\subsection{Support for simultaneous and sequential algorithms.} 
\label{app:sim_seq_alg}
\noindent\textbf{Methods.} We test sequential IL$\to$RL baselines comprising of the methods studied in Sec.~\ref{sec:exp-online-offline} and find that these IL$\to$RL methods significantly improve over using only IL or RL alone. Moreover, we train ADVISOR~\cite{advisor} which adaptively combines imitation and rewards-based losses to bridge the `imitation gap' between the expert and the agent. This is achieved via an auxiliary actor trained only via imitation loss, details of which can be found in~\cite{advisor}. In line with~\cite{advisor}, we find that ADVISOR's adaptive and parallel combination of IL and RL losses performs the best. Note that when referring to imitation in this study, the baselines learn from an expert policy.~\cite{advisor} lists three additional baselines where the agent learns from offline demonstrations.\\
\noindent\textbf{Task.} We deploy the above methods on the \textsc{Lava Crossing (LC) Corrupt (S15, N7)} task from~\cite{advisor}. This task is based on \textsc{LavaCrossing} in MiniGrid environment, where an agent, only based on its egocentric view, needs to navigate to a goal while avoiding rivers of lava. Lava in a cell indicates that the episode will end if the agent steps on it. \textsc{Corrupt} denotes that the (shortest-path) expert might provide corrupted supervision to the agent. Particularly, the expert policy becomes a random policy when the expert comes $\leq10$ steps from the goal. This tackles a realistic challenge of training agents which learn despite corruption or noise. \textsc{S15} indicates that the grid is $15\times15$ in size.~\textsc{N7} marks that there are a total of 7 horizontal and vertical lava rivers in the environment.

\subsection{Support for multi-agent systems.}
\label{app:multi_agent}
\begin{figure}
    \centering
    % {
    %   \phantomsubcaption\label{fig:}
    % }
    \includegraphics[width=0.5\textwidth]{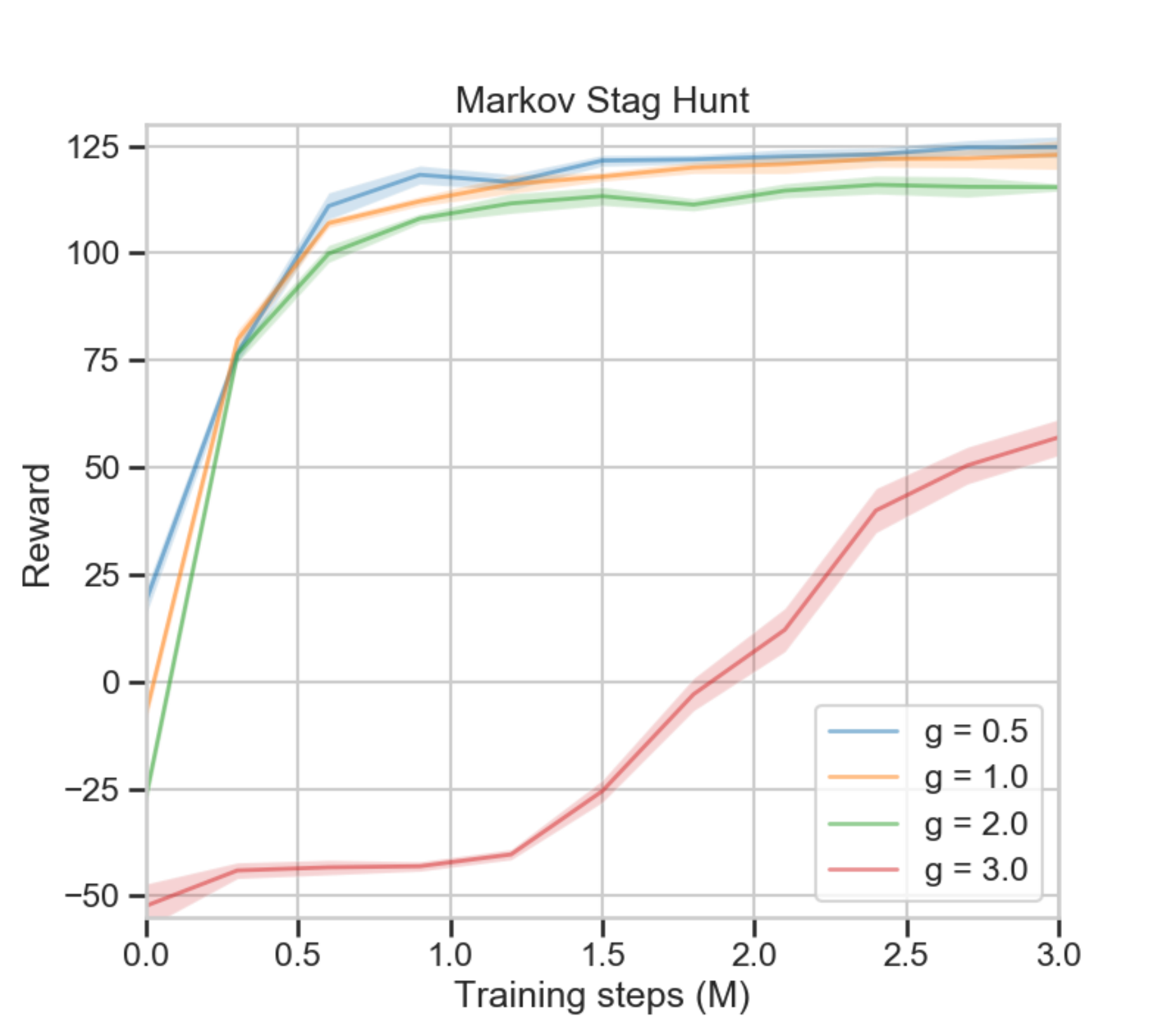}
\caption{\textbf{Testing results for the Markov Stag Hunt over training.} Here $g$ denotes the penalty, $g\in \{0.5,\ 1.0,\ 2.0,\ 3.0\}$, when only one is agent co-located with the Stag. The shaded regions indicate $95\%$ confidence intervals around the mean over $5$ different random seeds.}
    \label{fig:staghunt}
\end{figure}

We utilize MultiGrid\footnote{\url{https://github.com/ArnaudFickinger/gym-multigrid}} to reproduce the Markov Stag Hunt~\cite{peysakhovich2017prosocial}. MultiGrid, which is built within MiniGrid~\cite{minigrid}, is a grid-world environment developed for studying multi-agent reinforcement learning methods. On a $5 \times 5$ map, the agent and the Plant occupy $1 \times 1$ tiles, while the Stag occupies a $2 \times 2$ area. Following~\cite{peysakhovich2017prosocial}, we use the full map as the observation to the agents and this observation explicitly encodes entities' attributes with indices predefined by the MiniGrid. Our model includes an embedding layer with hidden size $8$ to encode the observation, a GRU with hidden size $512$ to process long-term memory, and a linear layer for actor-critic output (i.e., distribution over $4$ possible actions and value estimation). Both agents share the same embedding layer and GRU, while the linear layer's parameters are not shared.

We train the agents for $3$M steps (roughly $66.7K$ training episodes). We evaluate the learned model over $1000$ testing trajectories with $5$ different random seeds. As a result, we compute the average reward with its standard deviation over $5$ different random seeds. Test-time performance over training is shown in Fig.~\ref{fig:staghunt}. We observe that the agents converge to the payoff-dominant equilibrium when $g=\{0.5,\ 1.0,\ 2.0\}$ and risk-dominant equilibrium when $g=3.0$. In other words, the trained agents learn to cooperate and collect the Stag when $g=\{0.5,\ 1.0,\ 2.0\}$ but, when $g=3.0$, learn instead instead to individually focus on collecting the Plants. Thus, the payoff-dominant agents receive higher rewards than the risk-dominant agents.

% \section{Tutorials}\label{app:tutorials}

% In the below we include three tutorials from our \lib framework. To see the full documentation of \lib you will need to host our documentation locally on your machine. This is simple and instructions for doing so are outlined in the \texttt{DOCS-README.txt} file contained in our supplementary materials.

% \subsection{Running a simple experiment in MiniGrid}
% \centering\frame{\includegraphics[width=0.95\textwidth]{figs/minigrid_tutorial_0.1.pdf}}
% \includepdf[pages=2-,frame,width=\textwidth,pagecommand={}]{figs/minigrid_tutorial_0.1.pdf}

% \subsection{Running a point navigation experiment in RoboTHOR}
% \centering\frame{\includegraphics[width=\textwidth]{figs/pointnav_tutorial_0.1.pdf}}
% \includepdf[pages=2-,frame,width=\textwidth,pagecommand={}]{figs/pointnav_tutorial_0.1.pdf}

% \subsection{Running a point navigation experiment in iTHOR and Habitat}
% \centering\frame{\includegraphics[width=\textwidth]{figs/transfer_tutorial_0.1.pdf}}
% \includepdf[pages=2-,frame,width=\textwidth,pagecommand={}]{figs/transfer_tutorial_0.1.pdf}

\newpage
\bibliography{bibliography}

\end{document}